\documentclass[10pt,twocolumn,letterpaper]{article}

\usepackage{times}
\usepackage{epsfig}
\usepackage{graphicx}
\usepackage{amsmath}
\usepackage{amssymb}
\usepackage{changepage}
\usepackage{enumitem}
\usepackage{threeparttable}
\usepackage{subcaption}

\usepackage{booktabs, multirow} 
\usepackage{soul}
\usepackage[table]{xcolor} 
\usepackage{changepage,threeparttable} 
\usepackage{textcomp}
\usepackage{siunitx}

\def\wacvPaperID{181} 

\begin{document}

\title{Continual Learning with Neuron Activation Importance}

\author{Sohee Kim\qquad Seungkyu Lee\\
Kyunghee University, Department of Computer Engineering,\\
Yongin, Republic of Korea\\
{\tt\small soheekim@khu.ac.kr \qquad seungkhu@khu.ac.kr}

}

\maketitle

\begin{abstract}
Continual learning is a concept of online learning with multiple sequential tasks. One of the critical barriers of continual learning is that a network should learn a new task keeping the knowledge of old tasks without access to any data of the old tasks. In this paper, we propose a neuron activation importance-based regularization method for stable continual learning regardless of the order of tasks. 
We conduct comprehensive experiments on existing benchmark data sets to evaluate not just the stability and plasticity of our method with improved classification accuracy also the robustness of the performance along the changes of task order.
\end{abstract}

\section{Introduction}
Continual learning is a sequential learning scheme on multiple different tasks. New tasks do not necessarily consist of only existing classes of previous tasks nor statistically similar instances of existing classes. In challenging situations, new tasks may consist of mutually disjoint classes or existing classes with unseen types of instances in previous tasks. One of the main challenges is learning such new tasks without catastrophic forgetting existing knowledge of previous tasks.
Researchers have proposed diverse continual learning approaches to achieve both stability (remembering past tasks) and plasticity (adapting to new tasks) of their deep neural networks from sequential tasks of irregular composition of classes and varying characteristics of training instances.  
Since the training of a neural network is influenced more by recently and frequently observed data, the neural network forgets what it has learned in prior tasks without continuing access to them in the following tasks. 
A rigorous approach that maintains the knowledge of entire tasks may solve the problem while sacrificing computational cost, however, it is impractical with an undefined number of tasks in real applications of continual learning. 
Continual learning model has to adapt to a new task without access to some or entire classes of past tasks while it maintains acquired knowledge from the past tasks ~\cite{thrun1996learning}.
In addition, the continual learning model has to be evaluated with arbitrary order of tasks since the order of tasks is not able to be fixed nor predicted in real applications. The continual learning model is required to function consistently regardless of the order of tasks.

\begin{figure}[t!]
\begin{center}
\includegraphics[width= 1\linewidth]{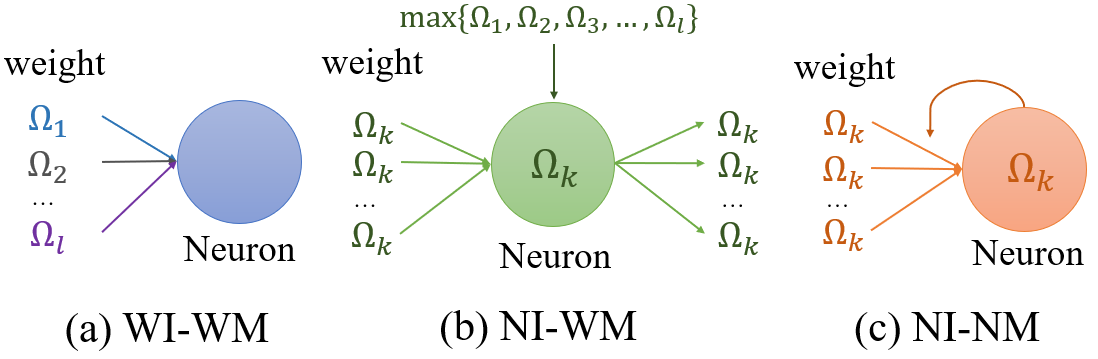}
\caption{Three different (Importance-Measurement) ways (a) WI-WM: Weight Importance ($\Omega_1 \sim \Omega_l$) by respective Weight Measurement (b) NI-WM: Neuron Importance ($\Omega_k$) by Weight Measurements. The maximum value of weight importance out of ($\Omega_1 \sim \Omega_l$) is assigned to neuron importance ($\Omega_k$). And then all weights connected to the neuron get the same importance of the neuron. (c) NI-NM: Neuron Importance ($\Omega_k$) by Neuron Measurement, where $l$ and $\Omega$ indicate weight index and importance of either weight or neuron respectively. The proposed method belongs to (c) NI-NM.}
\label{relate_concept}
\end{center}
\end{figure}

\begin{figure*}[t!]
  \begin{minipage}{.6\textwidth}
  \begin{subfigure}{.499\textwidth}
    \includegraphics[width=\linewidth]{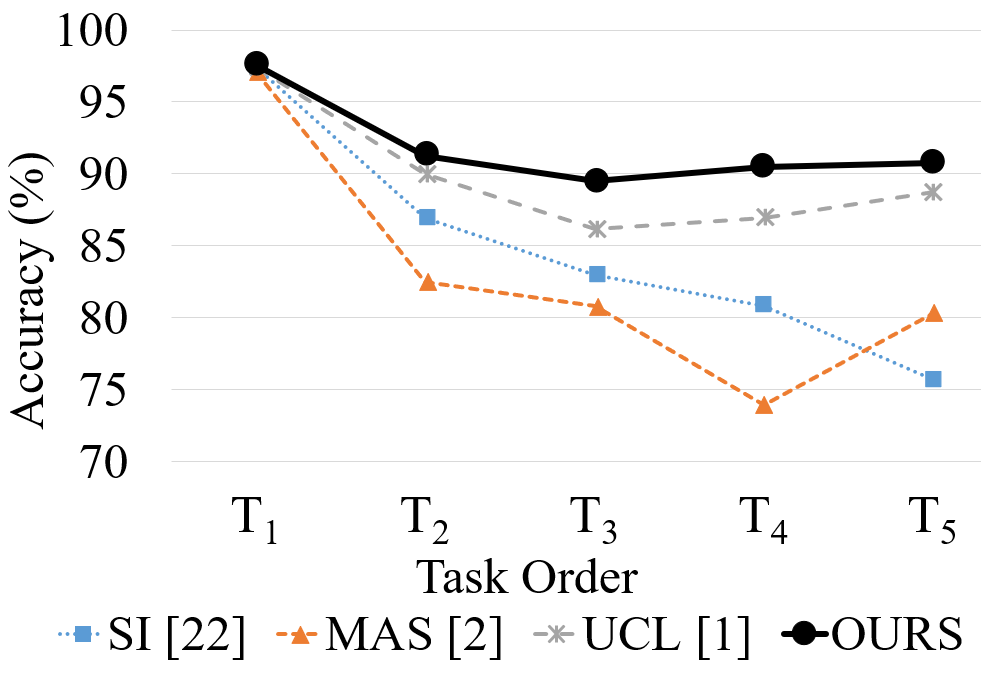}
    \caption{Task order: 1 $\rightarrow$ 2 $\rightarrow$ 3 $\rightarrow$ 4 $\rightarrow$ 5}
    \label{accu_diff1}
  \end{subfigure}
    \begin{subfigure}{.5\textwidth}
    \includegraphics[width=\linewidth]{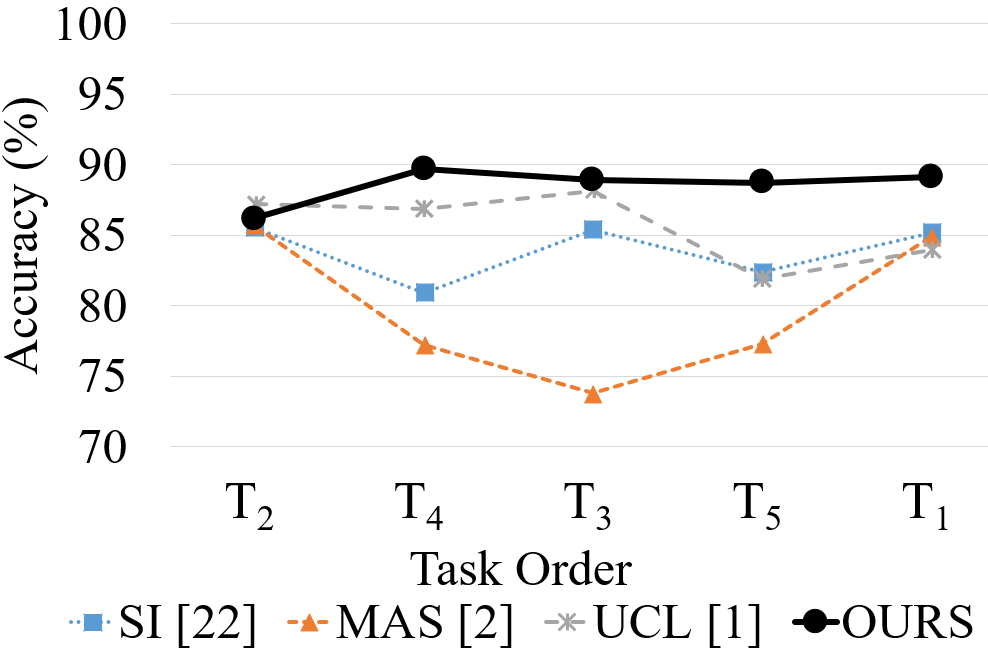}
    \caption{Task order: 2 $\rightarrow$ 4 $\rightarrow$ 3 $\rightarrow$ 5 $\rightarrow$ 1}
    \label{accu_diff2}
    \end{subfigure}
\caption{Classification accuracy of continual learning on Split Cifar10. SI~\cite{zenke2017continual}, MAS~\cite{aljundi2018memory} and UCL~\cite{ahn2019uncertainty} show critical changes in their performance as the order of tasks changes.}
\label{accu_diff}    
  \end{minipage} \quad 
  \begin{minipage}{.35\textwidth}
    \centering
    \resizebox{0.9\linewidth}{!}{
    \begin{tabular}{cccccc}
    \toprule
    \multicolumn{1}{c}{}&\multicolumn{5}{c}{  Absolute Task Order } \\ \cmidrule {2-6}
    Method        & T\textsubscript{1}     & T\textsubscript{2}     & T\textsubscript{3}     & T\textsubscript{4}     & T\textsubscript{5}     \\ \midrule
    SI~\cite{zenke2017continual}            & 11.9           & 5.9           & 2.4           & \textbf{1.5}           & 9.48           \\
    MAS~\cite{aljundi2018memory}           & 11.4           & 5.2           & 7.0           & 3.3           & 4.5            \\
    UCL~\cite{ahn2019uncertainty}           & \textbf{10.2}           & 3.1           & 2.0           & 5.0           & 4.7            \\ 
    \textbf{OURS}          & 11.4           & \textbf{1.5}           & \textbf{0.5}           & 1.7           & \textbf{1.6}  \\ \bottomrule
    \end{tabular}}%
    \captionof{table}{Performance disparity(\%) between Figure~\ref{accu_diff1} and ~\ref{accu_diff2} on Split CIFAR 10. "Absolute task order" represents the sequence of tasks that a model learns. (Additional explanation is discussed in Section~\ref{exp}.)}
    \label{performanceD}
  \end{minipage}
\end{figure*}


There are three major categories in prior continual learning approaches; 1) architecture modification of neural networks~\cite{yoon2017lifelong, sharif2014cnn, rusu2016progressive}, 2) rehearsal using sampled data from previous tasks~\cite{riemer2018learning, aljundi2019gradient}, and 3) regularization freezing significant weights of a model calculating the importance of weights or neurons~\cite{li2017learning, kirkpatrick2017overcoming, zenke2017continual, nguyen2017variational, aljundi2018memory, aljundi2018selfless, zeno2018task, ahn2019uncertainty, javed2019meta, jung2020adaptive}.
Most recent methods have tackled the problem with fundamental regularization approaches that utilize the weights of given networks to the fullest.
The basic idea of regularization approaches is to constrain essential weights of prior tasks not to change. In general, they alleviate catastrophic interference with a new task by imposing a penalty on the difference of weights between the prior tasks and the new task. The extent of the penalty is controlled by the significance of weights or neurons in solving a certain task using respective measurements. 
As Figure~\ref{relate_concept} illustrates, weight importance can be decided by three different (Importance-Measurement) ways. 

WI-WM (Weight Importance by Weight Measurement)~\cite{kirkpatrick2017overcoming, zenke2017continual, aljundi2018memory, nguyen2017variational, zeno2018task} calculates weight importance based on the measurement of the corresponding weight as described in Figure~\ref{relate_concept}a. 
Elastic weight consolidation (EWC)~\cite{kirkpatrick2017overcoming} estimates parameter importance using the diagonal of the Fisher information matrix equivalent to the second derivative of the loss. Synaptic intelligence (SI)~\cite{zenke2017continual} measures the importance of weights in an online manner by calculating each parameter's sensitivity to the loss change while it trains a network. When a certain parameter changes slightly during training batches but its contribution to the loss is high (i.e., rapid change of its gradient), the parameter is considered to be crucial and restricted not to be updated in future tasks. 
Unlike SI~\cite{zenke2017continual}, Memory aware synapses (MAS)~\cite{aljundi2018memory} assesses the contribution of each weight to the change of a learned function. It considers the gradient of outputs of a model with a mean square error loss. Gradient itself represents a change of outputs concerning the weights.
Variational Continual Learning (VCL)~\cite{nguyen2017variational}, a Bayesian neural network-based method, decides weight importance through variational inference. Bayesian Gradient Descent (BGD)~\cite{zeno2018task} finds posterior parameters (e.g., mean and variance) assuming that the posterior and the prior distributions are Gaussian. 


To mitigate the interference across multiple tasks in continual learning, weight importance-based approaches let each weight have its weight importance.
However, in the case of convolutional neural networks, since a convolutional filter makes one feature map that can be regarded as one neuron, those weights should have the same importance. Furthermore, those methods that consider the amount of change of weights~\cite{kirkpatrick2017overcoming, zenke2017continual, aljundi2018memory} are impossible to reinitialize weights at each training of a new task, which possibly decreases the plasticity of the network. (Additional explanation of weight re-initialization is discussed in section~\ref{weight_reinit}.)

NI-WM (Neuron Importance by Weight Measurement) calculates neuron importance based on the measurement of all weights.
Weight importance is redefined as the importance of its connected neuron~\cite{ahn2019uncertainty}. Uncertainty-regularized Continual Learning (UCL)~\cite{ahn2019uncertainty} measures weight importance by its uncertainty indicating the variance of weight distribution. It claims that the distribution of essential weights for past tasks has low variance, and such stable weights during training a task are regarded as important weights not to forget. As illustrated in Figure~\ref{relate_concept}b, it suggests neuron-based importance in neural networks. 
The smallest variance value (maximum importance) among the weights incoming to and outgoing from a corresponding neuron decides the importance of the neuron, and then the importance of all those weights is updated as the neuron importance.

\begin{figure}[b!]
\begin{center}
\includegraphics[width=1\linewidth]{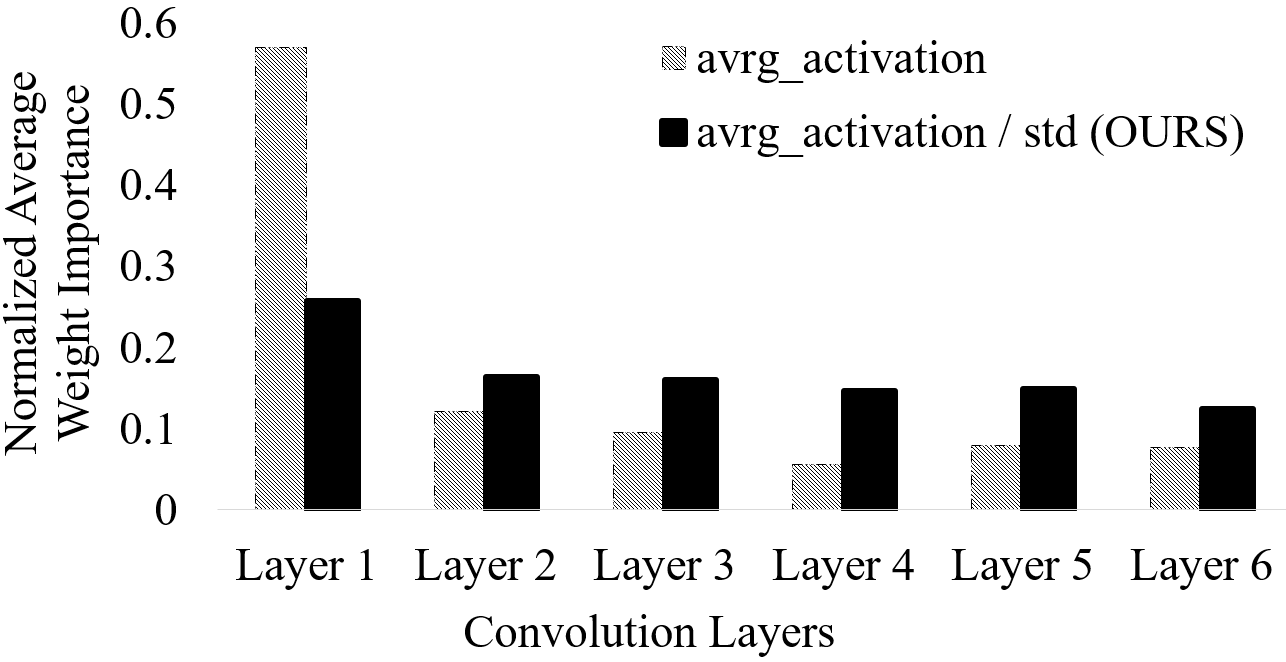}
\caption{Normalized Weight importance distribution of each convolution layer. To show the proportion of the average value of weight importance among layers, we normalize the values to sum 1. This is based on the first task of Split CIFAR 10 (task order: 3-1-2-4-5).}
\label{WID}
\end{center}
\end{figure}

NI-NM (Neuron Importance by Neuron Measurement) calculates neuron importance based on the measurement of the corresponding neuron~\cite{jung2020adaptive}. Weight importance is defined as the importance of its connected neuron.

~\cite{jung2020adaptive} exploits proximal gradient descents using a neuron importance. 
Its neuron importance depends on the average activation value. Activation value itself is a measurement of neuron importance, and weights connected to the neuron get identical weight importance. 

One critical observation in prior experimental evaluations of existing continual learning methods is that the accuracy of each task significantly changes when the order of tasks is changed. 
As discussed in ~\cite{yoon2019scalable}, proposing a continual learning method robust to the order of tasks is another critical aspect. 
Therefore, performance evaluation with fixed task order does not coincide with the fundamental aim of continual learning where no dedicated order of tasks is given in reality. 
Figure~\ref{accu_diff} shows sample test results of state-of-the-art continual learning methods compared to our proposed method. As summarized in Table~\ref{performanceD}, classification accuracy values of prior methods fluctuate as the order of tasks changes(from Figure~\ref{accu_diff1} to Figure~\ref{accu_diff2}). 
%

In this work, we propose a regularization approach for continual learning assigning neuron importance by the measurement of average neuron activation. 
As Figure~\ref{WID} describes, we balance neuron importance distribution among layers based on the average neuron activation divided by standard deviation, which is critical to performance consistency along the changes of task order.
We assign calculated neuron importance to all weights of incoming edges connected to the neuron.

A Neuron with high activation to the majority of instances is defined as an essential neuron.
We freeze essential neurons by freezing the weights of all connected incoming edges (essential weights) during the learning of a new task so that our model remembers past tasks. 
We propose to evaluate the robustness to the order of tasks in a comprehensive manner in which we evaluate the average and standard deviation of classification accuracy with multiple sets of randomly shuffled tasks.

Our approach remembers past tasks robustly compared to recent regularization methods~\cite{zenke2017continual, nguyen2017variational, aljundi2018memory,ahn2019uncertainty}.
To measure performance fluctuation along the change of task order, we evaluate our method with numerous shuffled orders. We quantitatively evaluate our classification performance based on a measure of interference from past tasks on MNIST~\cite{lecun1998gradient, goodfellow2013empirical}, CIFAR10, CIFAR100~\cite{krizhevsky2009learning} and Tiny ImageNet~\cite{deng2020imagenet} data sets. 

Key contributions of our work include 1) a simple but intuitive and effective continual learning method introducing activation based neuron importance, 2) a comprehensive experimental evaluation framework on existing benchmark data sets to evaluate not just the final accuracy of continual learning also the robustness of the accuracy along the changes of the order of tasks. Based on the evaluation framework, existing state-of-the-art methods and the proposed method are evaluated.

\section{Proposed Method}
\subsection{Neuron Importance by Average Neuron Activation}
\begin{figure}[t!]
\begin{center}
\includegraphics[width=1\linewidth]{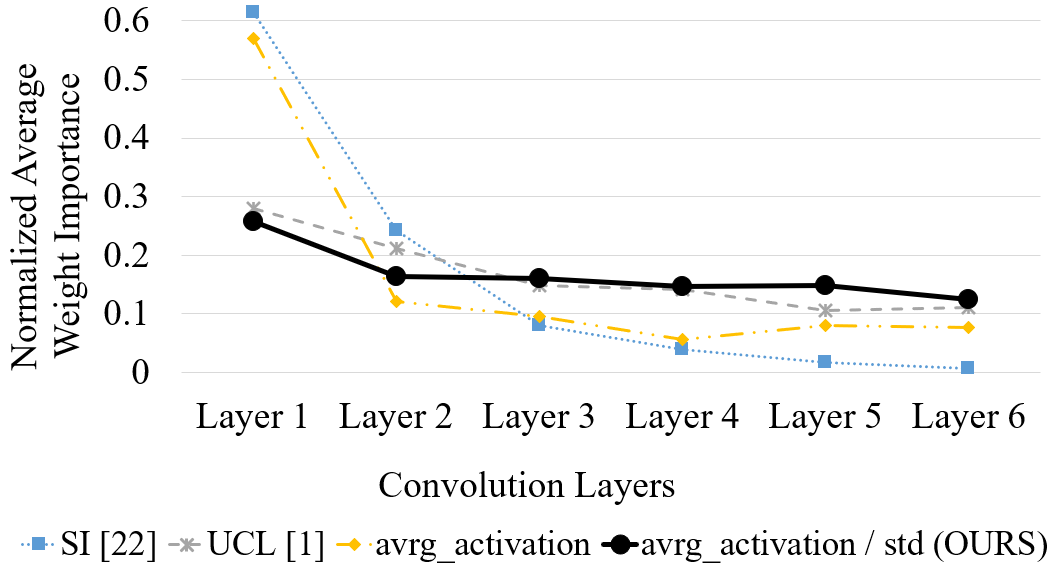}
\caption{Normalized Weight importance distribution of each convolution layer. To show the proportion of the average value of weight importance among layers, we normalize the values to sum 1. Our method relaxes the tendency to excessively consolidate weights of earlier layers. This is based on the first task of Split CIFAR 10 (task order: 3-1-2-4-5).}
\label{omega_dist_graph}
\end{center}
\end{figure}

The proposed method extracts neuron importance based on the average activation value of all instances.
And then the neuron importance is assigned to all weights of incoming edges connected to the neuron.
In convolutional neural networks, activation value of a neuron corresponds to the average value of one feature map (i.e., global average pooling value). The average activation value of neuron corresponds to the average of global average pooling value.
The average activation values at each layer are independently calculated but are considered together. In other words, the individual average activation values represent the importance of each neuron of a whole model.
However, encoded features at each layer describe different aspects of an input image and, as a result, the average activation values at each layer should not be evaluated together.
Therefore, the average activation value is not able to fully represent the characteristics of the essential neuron. 
Besides, in convolution neural networks, the absolute magnitude of average activation value (i.e., the average of global average pooling value) varies along the location of layer: in high-level feature maps, the portion of activated area decreases. 
Due to the difference in absolute average activation values across the layers, weights of earlier layers tend to be considered more essential as Figure~\ref{omega_dist_graph} shows. If the average activation value is used as neuron importance, networks will prefer to keep the weights of earlier layers.


Instead, we propose to use layer-wise average activation divided by the respective standard deviation for neuron importance measurement. 
Compared to the average activation-based neuron importance~\cite{jung2020adaptive}, ours prevents earlier layers from getting excessive importance compared to other layers, which, in turn, prevents a network from vulnerable to changing the order of tasks in terms of forgetting past tasks.  
Figure~\ref{omega_dist_graph} shows normalized average weight importance of each layer(total 6 layers). Prior average activation based regularization term assigns around 57\% of total importance to layer 1(57\%, 12\%, 10\%, 6\%, 8\%, 8\%, respectively for the 6 layers). On the other hand, our proposed regularization loss term assigns 26\% of total importance to layer 1. Furthermore, our method avoids assigning excessive importance to certain layer(26\%, 16\%, 16\%, 15\%, 15\%, 12\%).

Then, why this improves the continual learning performance regardless of task order?
In prior works, more weights of lower layers tend to be frozen in earlier tasks that eliminate the chance of upcoming tasks to build new low-level feature sets. Only a new task that is fortunately able to rebuild higher-layer features based on the frozen lower layer weights from previous tasks could survive. On the other hand, ours keeps the balance of frozen weights in all layers securing more freedom of feature descriptions for new tasks in both lower and higher layers. Indeed, lower layer features such as edges are not class (task) dependent features.
Therefore, excessively freezing lower layer features is not preferable in continual learning.
Even though tasks change, a new task may find alternative low-level features that have high similarity with them of past tasks, as discussed in ~\cite{kornblith2019similarity}.
In order to encode such relation, we propose to use the average and standard deviation of neuron activation values at each layer. 
Our loss function is described as follows.
\begin{equation}
\begin{aligned}
L_{t} = \tilde{L_{t}} + \alpha \sum\limits_{l} \Omega_{k}^{t}(w_l^{t-1}-w_l^t)^2, 
\end{aligned}
\end{equation}
where $\tilde{L_{t}}$ is loss of current task (e.g., cross entropy loss), $t$ is task index, $l$ is weight index, and $\Omega_{k}^{t}$ indicates $k^{th}$ neuron importance. $\alpha$ is a strength parameter to control the amount of weights consolidation. Neuron importance is defined as follows.

\begin{equation}
\begin{gathered}
\Omega^{t}_{k} =  \frac{\frac{1}{N_{t}}\sum\limits_{{i = 1}}^{N_t}{f_{k}(x^{(t)}_{i})}}{\sigma + \epsilon},\\
\sigma = \sqrt{\frac{\sum\limits_{{i = 1}}^{N_t}{\{{f_{k}(x^{(t)}_{i})-\frac{1}{N_{t}}\sum\limits_{{i = 1}}^{N_t}{f_{k}(x^{(t)}_{i})}}\}^2}}{N_t}},
\end{gathered}
\end{equation}
where $N_t$ is the number of instances, $x$ is input, $k$ is neuron index, $f_{k}(\cdot)$ is activation value (global average value, in the case of convolution neural network), and $i$ is instance index. We introduce $\epsilon$ to prevent the numerator from being zero when the standard deviation becomes zero.
Proposed method considers the variation of average activation value among instances and the differences of average activation value among different layers. 
It encourages freezing more weights of later layers than earlier layers which are more likely to describe given task-specific features.

Our experiments(Table~\ref{cifar10_forget} in Section~\ref{e_cifar10}) show that prior methods tend to forget past tasks in learning new tasks. 
In the prior methods, weights of later layers are more likely to change than weights of earlier layers during learning a new task.

In general, if the essential weights of later layers of previous tasks change, the network forgets past tasks and hardly recovers previous task-specific features. 
On the other hand, even though weights of earlier layers of previous tasks change, there are other chances to recover general low-level features which are shared with following new tasks.
Since our method puts relatively more constraints on the weights of task-specific features not to change than the prior methods(Figure~\ref{WID}), our method forgets past tasks less showing stable performance along the change in the order of tasks. 
 
\subsection{Weight Re-initialization for Better Plasticity}
\label{weight_reinit}
In continual learning, networks have to not only avoid catastrophic forgetting but also learn new tasks. According to the extent of difference in optimal classification feature space of different tasks, optimized feature space in the previous task might be significantly changed with a new task. In the learning of a new task, we can let the model start either from random weights or from optimized weights with previous tasks.
Even though the optimized weights on previous tasks can be considered as a set of random weights for a new task, we avoid a situation where the optimized weights for one task work as a local optimal for another similar task that may hinder new training from obtaining new optimal weights through weight re-initialization.
The situation can be explained with $\Omega_{k}(w_k^{t-1}-w_k^t)^2$ term in the loss function of our network. During the learning of a new task, the network is informed of past tasks by $\Omega_{k}(w_k^{t-1}-w_k^t)^2$ term which lets the network maintain essential weights of the past tasks assigning high $\Omega_{k}$ values. In other words, $\Omega_{k}(w_k^{t-1}-w_k^t)^2$ delivers the knowledge of previous tasks. Whatever the magnitude of $\Omega_{k}$ is, however, $\Omega_{k}(w_k^{t-1}-w_k^t)^2$ term is ignored if $w_k^{t-1}$ almost equals to $w_k{t}$ already in the initial epoch of the training of a new task, which prevents the network from learning a new task.
This situation is alleviated by weight re-initialization that allows the value of $\Omega_{k}(w_k^{t-1}-w_k^t)^2$ to be high enough regardless of the magnitude of $\Omega_{k}$ in the training of a new task. In this case, still the knowledge of previous tasks will be delivered by $\Omega_{k}$ and affect the training of a new task.

\section{Experimental Evaluations}
\label{exp}
We perform experimental evaluations of our method compared to existing state-of-the-art methods for continual learning on several benchmark data sets; Split and permuted MNIST~\cite{lecun1998gradient, goodfellow2013empirical}, and incrementally learning classes of CIFAR10, CIFAR100~\cite{krizhevsky2009learning} and Tiny ImageNet~\cite{deng2020imagenet}.
We set hyper-parameters of other existing approaches based on the description in~\cite{ahn2019uncertainty} which has tested existing approaches with different hyper-parameters to find their best performance. We train all different tasks with a batch size of 256 and Adam~\cite{kingma2014adam} using the same learning rate (0.001).
For the Split CIFAR tasks and Split Tiny ImageNet, as aforementioned, we perform the evaluation multiple times shuffling the order of tasks randomly to evaluate the robustness to task orders.
We test with all 120, 200, and 50 random orders for Split CIFAR10, Split CIFAR10-100 and Split Tiny ImageNet respectively.
To minimize statistical fluctuations of accuracy, each combination of task sequences is repeated three times.
\begin{figure}[t!]
\begin{center}
\includegraphics[width= 1\linewidth]{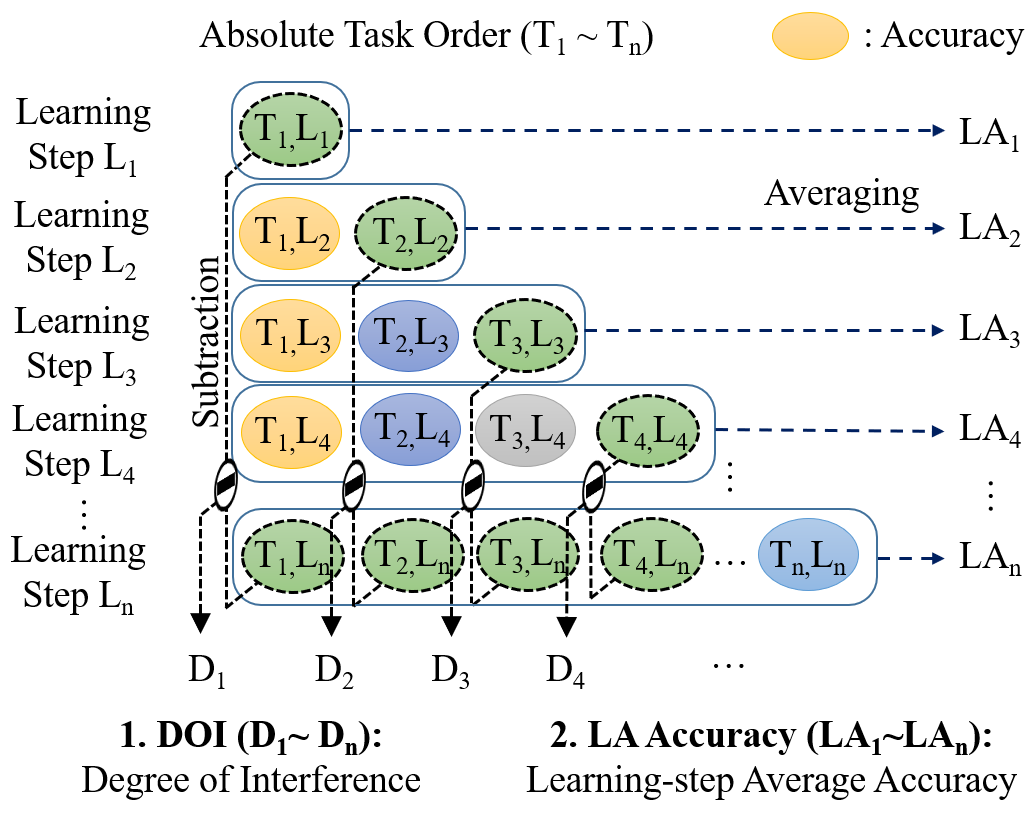}
\caption{Evaluation Metrics: 
DOI(Degree of Interference) and LA Accuracy of task. T, L and n stands for task, learning step, and the number of tasks respectively.}
\label{eval}
\end{center}
\end{figure}

As described in Figure~\ref{eval}, we define several evaluation metrics. "Absolute task order" indicates the sequence of tasks that a model learns. For instance, task 1 stands for the first task that a model learns no matter which classes comprise the task. "Learning step-wise average accuracy(LA Accuracy)" represents the accuracy of each learning step averaged through the whole tasks involved. (i.e., $LA_k = Average(L_k)$).
"Degree of interference(DOI)" indicates the decreased extent of accuracy of each task after all learning steps are conducted. It is calculated by $(T_k, L_k) - (T_k, L_n)$.
When we report the performance of randomly shuffled order experiment, we respectively average LA accuracy and DOI of randomly shuffled ordered test.

\subsection{MNIST}
\begin{figure}[b!]
\begin{center}
\includegraphics[width= 1\linewidth]{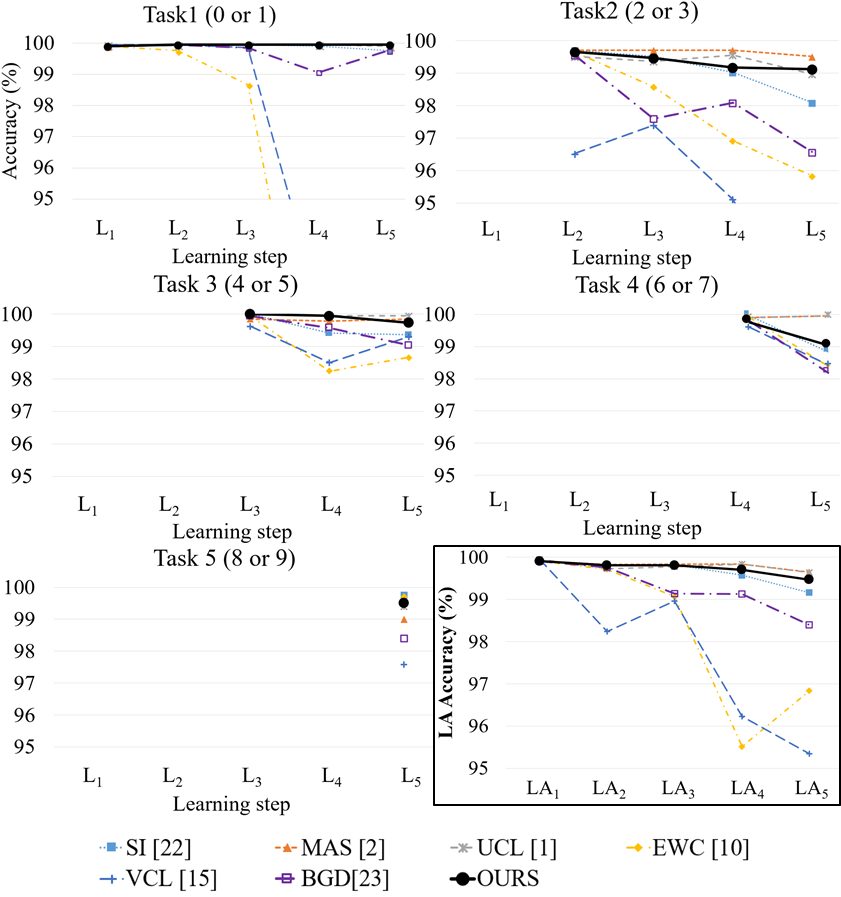}
\caption{Results on Split MNIST benchmark. Here, VCL indicates VCL(without coreset)\cite{nguyen2017variational}.}
\label{split_mnist}
\end{center}
\end{figure}

We first evaluate our algorithm on a Split MNIST benchmark. In this experiment, two sequential classes compose each task (total 5 tasks). We use multi-headed and multi-layer perceptrons with two hidden layers with 400 ReLU activations. Each task has its output layer with two outputs and Softmax. We train our network for 40 epochs with $\alpha$ = 0.0045.
In Figure~\ref{split_mnist}, we compare the accuracy of each task for at every learning step (column-wise comparison in Figure~\ref{eval}) and LA accuracy.
MAS~\cite{aljundi2018memory} outperforms all other baselines reaching 99.81\% while ours achieves 99.7\%. However, the accuracy is almost saturated due to the low complexity of the data. 

We also evaluate methods on permuted MNIST data set. Our model used in this evaluation is MLP which consists of two hidden layers with 400 ReLUs each and one output layer with Softmax. The network is trained for 20 epochs with $\lambda$ = 0.005. Also, to normalize the range of activation value, ReLU is applied to the output layer additionally when computing neuron importance $\Omega_{k}$. Our algorithm (95.21\%) outperforms MAS~\cite{aljundi2018memory} (94.70\%), EWC~\cite{kirkpatrick2017overcoming} (82.45\%) and VCL(without coreset)~\cite{nguyen2017variational} (89.76\%) and on the other hand, UCL~\cite{ahn2019uncertainty} (96.72\%), SI~\cite{zenke2017continual} (96.39\%) and BGD~\cite{zeno2018task} (96.168\%) show better results. However, most results on this data set achieve almost saturated accuracy.

\subsection{Split CIFAR10}
\label{e_cifar10}
We test our method on a Split CIFAR10 benchmark. In this experiment, two sequential classes compose each task (total 5 tasks). Evaluation on Split CIFAR10 data set is based on the multi-headed network with six convolution layers and two fully connected layers where the output layer is different for each task. We train our network for 100 epochs with $\alpha = 0.7$. The order of 5 tasks that comprise CIFAR10 is randomly shuffled (total 120 random orders).

\begin{figure}[b!]
\begin{center}
\includegraphics[width= \linewidth]{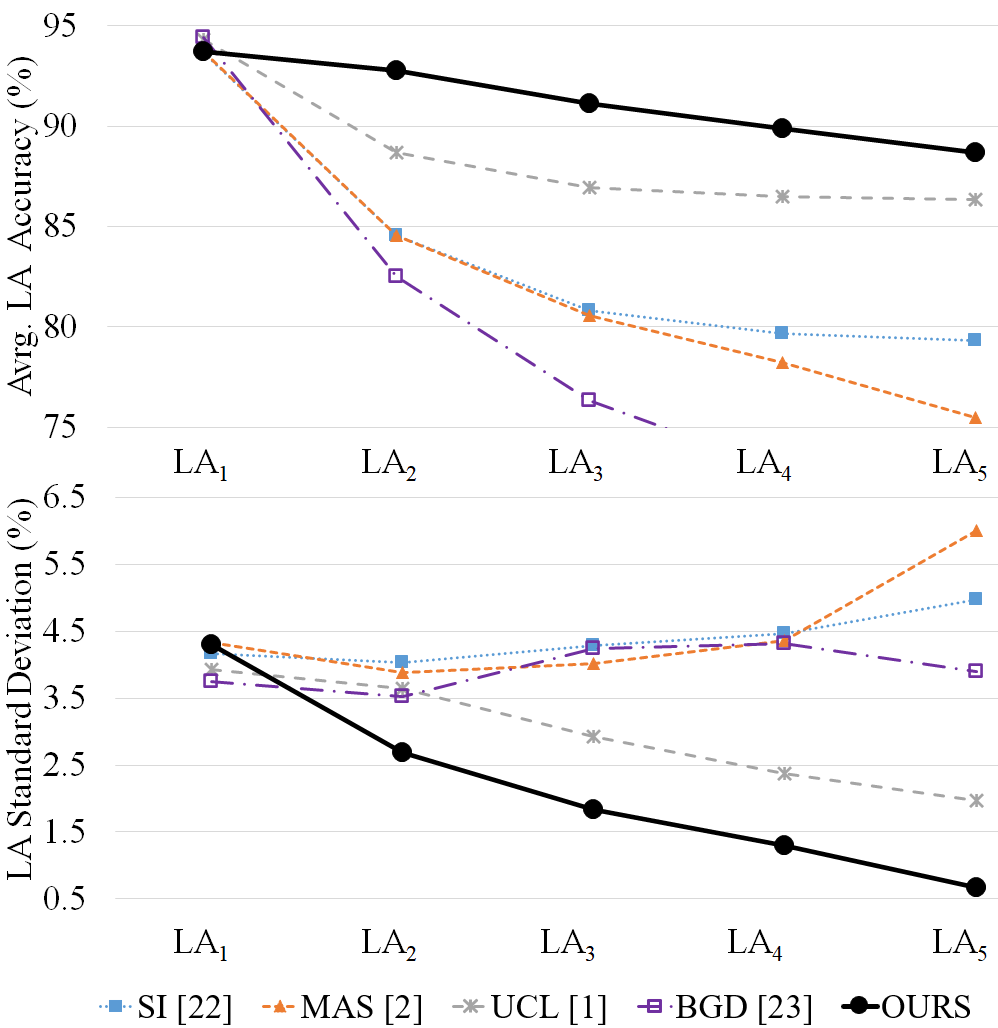}
\caption{Average LA Accuracy and its std. of Split CIFAR10 benchmark.}
\label{cifar10}
\end{center}
\end{figure}
\begin{table}[b!]
\centering
\resizebox{\linewidth}{!}{%
\begin{tabular}{cccccc}
\toprule
\multicolumn{1}{c}{}&\multicolumn{4}{c}{ Average DOI of Absolute Task Order } \\ \cmidrule {2-5}
Method & D\textsubscript{1}       & D\textsubscript{2}       & D\textsubscript{3}       & D\textsubscript{4}       \\ \midrule
SI~\cite{zenke2017continual}     & 28.05(\textpm11.7) & 20.00(\textpm7.4) & 15.51(\textpm8.2) & 8.68(\textpm5.8) \\
MAS~\cite{aljundi2018memory}    & 33.59(\textpm11.7) & 27.37(\textpm11.3) & 19.15(\textpm10.6) & 11.45(\textpm6.7) \\
UCL~\cite{ahn2019uncertainty}    & 11.36(\textpm5.8) & 8.56(\textpm3.6) & 5.94(\textpm3.0)  & 3.55(\textpm6.5) \\
BGD~\cite{zeno2018task}    & 39.06(\textpm10.1) & 34.83(\textpm8.5) & 29.19(\textpm8.8) & 19.71(\textpm2.1) \\
 \textbf{OURS}   &  \textbf{1.44(\textpm1.1)} &  \textbf{1.59(\textpm1.2)} &  \textbf{1.18(\textpm0.8)}  &  \textbf{0.70(\textpm0.7)}\\ \bottomrule
\end{tabular}%
}
\caption{Average DOI(Degree of interference) and its std.(\%) on Split CIFAR10. Note that proposed method forgets past tasks less regardless of the order of tasks.}
\label{cifar10_forget}
\end{table}
As Figure~\ref{cifar10} describes, our method overall outperforms all other methods with large margins. Also, the standard deviation graph shows that our algorithm is more robust to the order of tasks.

As Table~\ref{cifar10_forget} shows, proposed method shows better stability in the order of tasks and also has a low degree of forgetting. In our method, average degraded degree of performance is lowest as $1.23\%$, whereas SI~\cite{zenke2017continual} is $18.06\%$, UCL~\cite{ahn2019uncertainty} is $7.35\%$, MAS~\cite{aljundi2018memory} is $22.89\%$, and BGD~\cite{zeno2018task} is $30.7\%$.

\subsubsection{Ablation study}
To verify the effect of weight re-initialization for the learning of new tasks, we compare performance of ours and UCL~\cite{ahn2019uncertainty} with those without weight re-initialization.
\begin{table}[t!]
\centering
\resizebox{0.9\linewidth}{!}{%
\begin{tabular}{cccccc}
\toprule
\multicolumn{1}{c}{}&\multicolumn{5}{c}{ Task Order } \\ \cmidrule {2-6}
Method  & T\textsubscript{5} & T\textsubscript{4} & T\textsubscript{3} & T\textsubscript{2} & T\textsubscript{1}     \\ \midrule
UCL~\cite{ahn2019uncertainty}  & 0      & -0.425 & -0.9   & 4.93   & 6.38   \\
\textbf{OURS} & -1     & 15.7   & 21.44  & 18.62  & 21.44 \\ \bottomrule
\end{tabular}%
}
\caption{Performance difference(\%) =  (accuracy with weight re-initialization) - (accuracy without weight re-initialization). Note that the task order is fixed.}
\label{weight_reinit}
\end{table}
As Table~\ref{weight_reinit} indicates, accuracy increases in both methods when weight re-initialization is applied. It suggests that weight re-initialization encourages better plasticity. Note that several weight importance based methods~\cite{kirkpatrick2017overcoming,zenke2017continual,aljundi2018memory} cannot employ weight re-initialization since they consider the amount of weight changes in the methods.

\subsection{Split CIFAR10-100}
\begin{figure}[t!]
\begin{center}
\includegraphics[width= 1\linewidth]{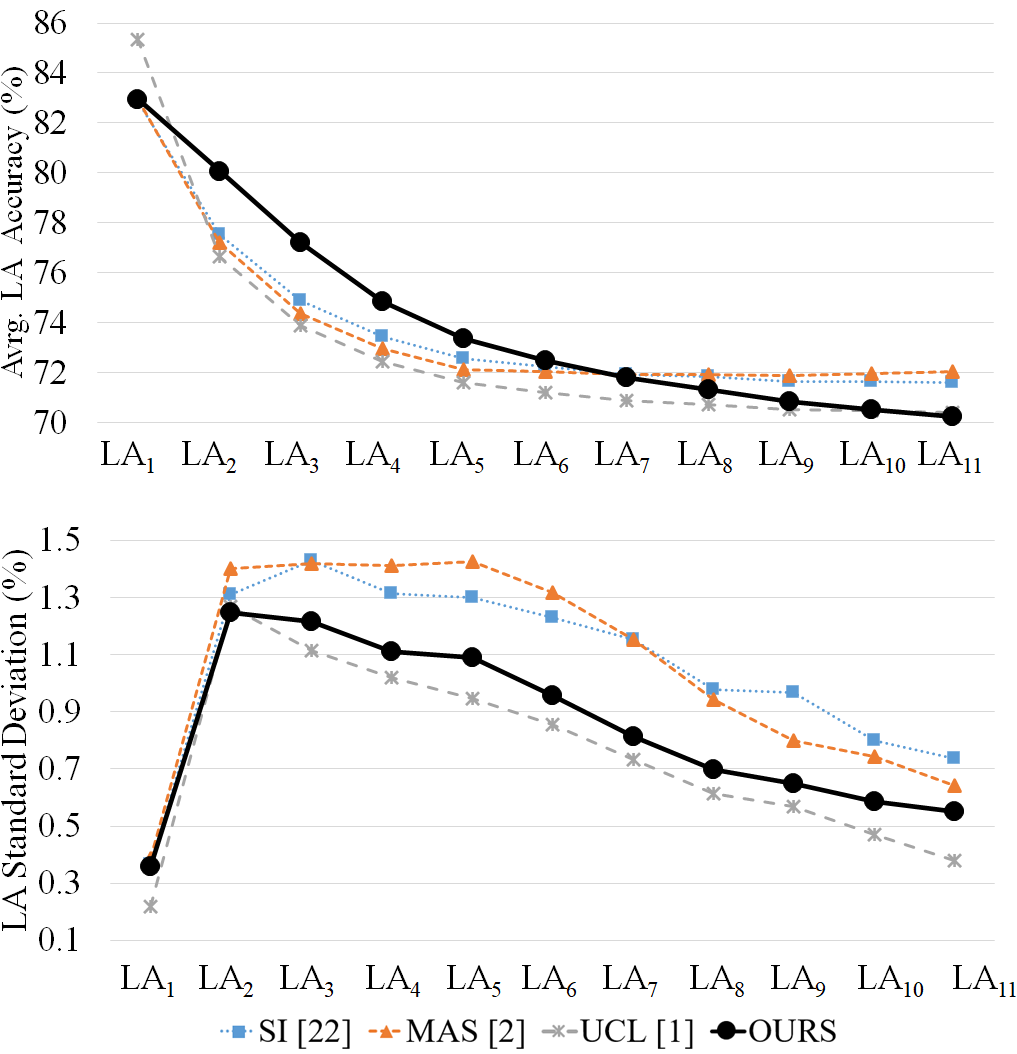}
\caption{Average LA Accuracy and its std. of CIFAR10-100 benchmark.}
\label{cifar100}
\end{center}
\end{figure}

\begin{table*}[t!]
\centering
\resizebox{\linewidth}{!}{%
\begin{tabular}{ccccccccccc}
\toprule
\multicolumn{1}{c}{}&\multicolumn{10}{c}{  Average DOI of Absolute Task Order } \\ \cmidrule {2-11}
Method        & D\textsubscript{1}               & D\textsubscript{2}              & D\textsubscript{3}                & D\textsubscript{4}             & D\textsubscript{5}                & D\textsubscript{6}              & D\textsubscript{7}                & D\textsubscript{8}             & D\textsubscript{9}                & D\textsubscript{10}               \\ \midrule
SI~\cite{zenke2017continual}            & 5.85(\textpm0.9)          & 7.37(\textpm2.3)        & 6.58(\textpm2.1)          & 5.87(\textpm2.0)       & 5.57(\textpm1.7)          & 5.09(\textpm1.7)        & 4.45(\textpm1.5)          & 3.97(\textpm1.3)       & 3.28(\textpm1.2)          & 2.17(\textpm1.0)          \\
MAS~\cite{aljundi2018memory}           & 9.32(\textpm1.4)         & 9.18(\textpm2.8)         & 8.19(\textpm2.2)          & 7.39(\textpm2.1)       & 6.65(\textpm1.9)          & 6.17(\textpm1.9)        & 5.30(\textpm1.6)          & 4.64(\textpm1.4)       & 3.70(\textpm1.2)          & 2.50(\textpm1.0)          \\
UCL~\cite{ahn2019uncertainty}           & 3.74(\textpm0.4)         & \textbf{1.20(\textpm1.2)}         & 1.31(\textpm1.1)          & 0.98(\textpm1.0)       & \textbf{0.78(\textpm0.9)}          & \textbf{0.68(\textpm0.8)}        & 0.64(\textpm0.7)          & 0.59(\textpm0.7)       & \textbf{0.37(\textpm0.5)}          & 0.27(\textpm0.4)          \\
\textbf{OURS} & \textbf{2.03(\textpm0.4)} & 2.08(\textpm0.9) & \textbf{1.26(\textpm0.9)} & \textbf{0.94(\textpm0
8)} & 0.84(\textpm0.8) & 0.77(\textpm0.7) & \textbf{0.61(\textpm0.7)} & \textbf{0.59(\textpm0.6)} & 0.47(\textpm0.5) & \textbf{0.24(\textpm0.4)} \\ \bottomrule
\end{tabular}
}
\caption{Average DOI(Degree of interference) and its std.(\%) on Split CIFAR10-100. Note that proposed method forgets past tasks less regardless of the order of tasks.}
\label{cifar100_forget}
\end{table*}

We evaluate our method on Split CIFAR10-100 benchmark where each task has 10 consecutive classes (total 11 tasks). We use the same multi-headed setup as in the case of Split CIFAR10. We train our network for 100 epochs with $\alpha = 0.5$. We fix task 1 as CIFAR10 due to the difference in the size of data set between CIFAR10 and CIFAR100. The order of remaining tasks that consist of CIFAR100 is randomly shuffled (total 200 random orders).

Our method shows better stability showing the best accuracy values in old tasks. On the other hand, previous methods seem to prefer to be better with recent new tasks proving that our importance based continual learning is working appropriately. Indeed, as Figure~\ref{cifar100} and Table~\ref{cifar100_forget} represent, SI~\cite{zenke2017continual} and MAS~\cite{aljundi2018memory} seem that they learn new tasks very well forgetting what they have learned before.
\begin{figure}[t!]
 \begin{subfigure}{\linewidth}
    \centering
    \includegraphics[width=\linewidth]{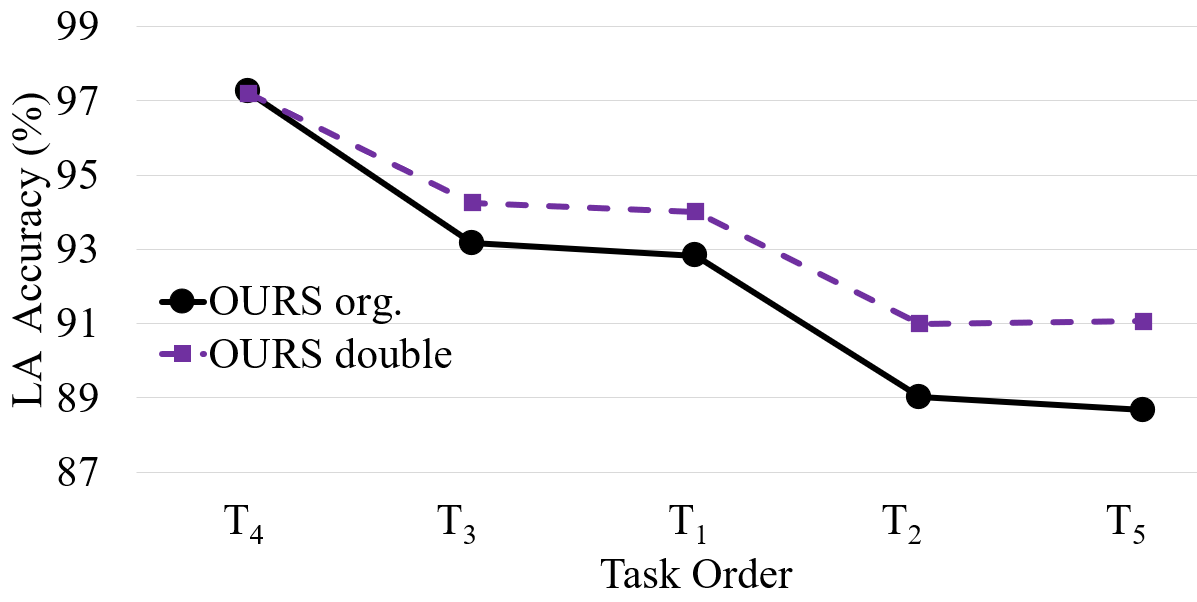}
    \caption{LA accuracy of Split Cifar10} \label{double_la}
 \end{subfigure}%
    \vskip\baselineskip
 \begin{subfigure}{\linewidth}
    \centering
    \includegraphics[width=\linewidth]{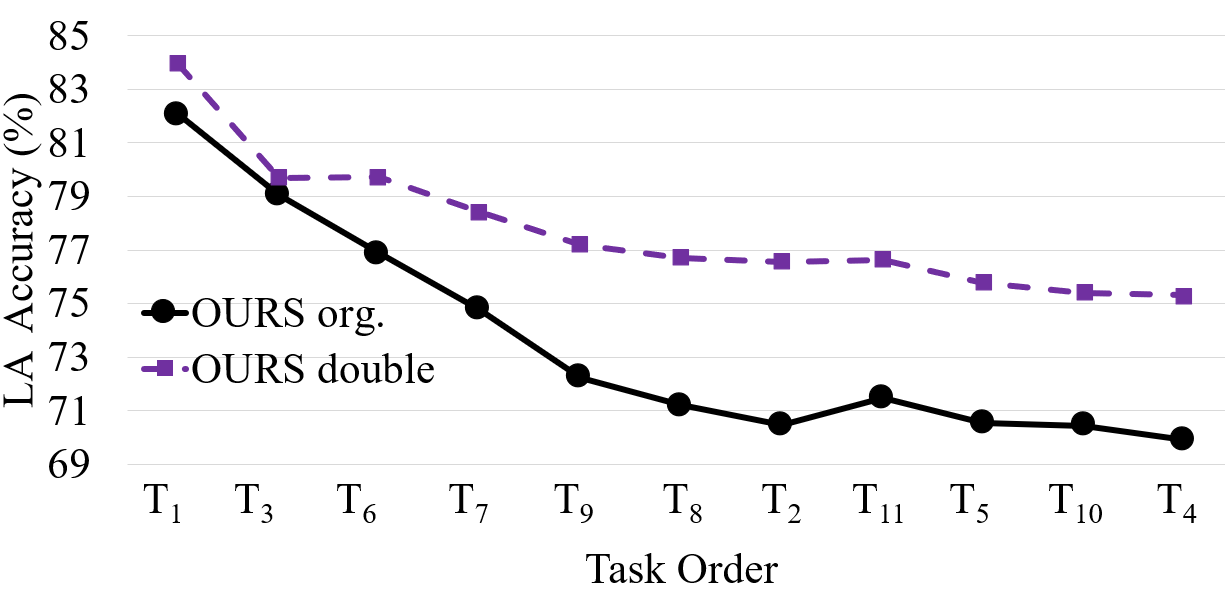}
    \caption{LA accuracy of Split Cifar10-100} \label{fig20:(a)}
 \end{subfigure}%
    \caption{The performance on Split CIFAR10 and CIFAR10-100 with doubled channel. Accuracy increases when we use a doubled channel network. Note that the task order is fixed.}
    \label{doubbled_cifar10_cifar100}
\end{figure}

Since all incoming weights are tied to the neuron in our method, the higher number of weights to be consolidated during training new tasks causes lower accuracy of final task. 
In practice, the decrease of plasticity in our method can be addressed by using a larger network (e.g., the larger number of channels).
We test the performance with a network of a doubled number of channels (256 to 512).
Figure~\ref{doubbled_cifar10_cifar100} shows that our network with doubled number of channels has improved accuracy keeping its stability and better plasticity.

Table~\ref{cifar100_forget} shows that our method obtains lowest average degraded degree of performance $0.98\%$ compared to SI~\cite{zenke2017continual}, MAS~\cite{aljundi2018memory}, UCL~\cite{ahn2019uncertainty} achieving $5.02\%$, $6.3\%$, $1.06\%$ respectively.
Also, the proposed method shows the lowest standard deviation of DOI, which indicates that our method is robust to the interference from various combinations of tasks.
\begin{figure}[t!]
\begin{center}
\includegraphics[width= 1\linewidth]{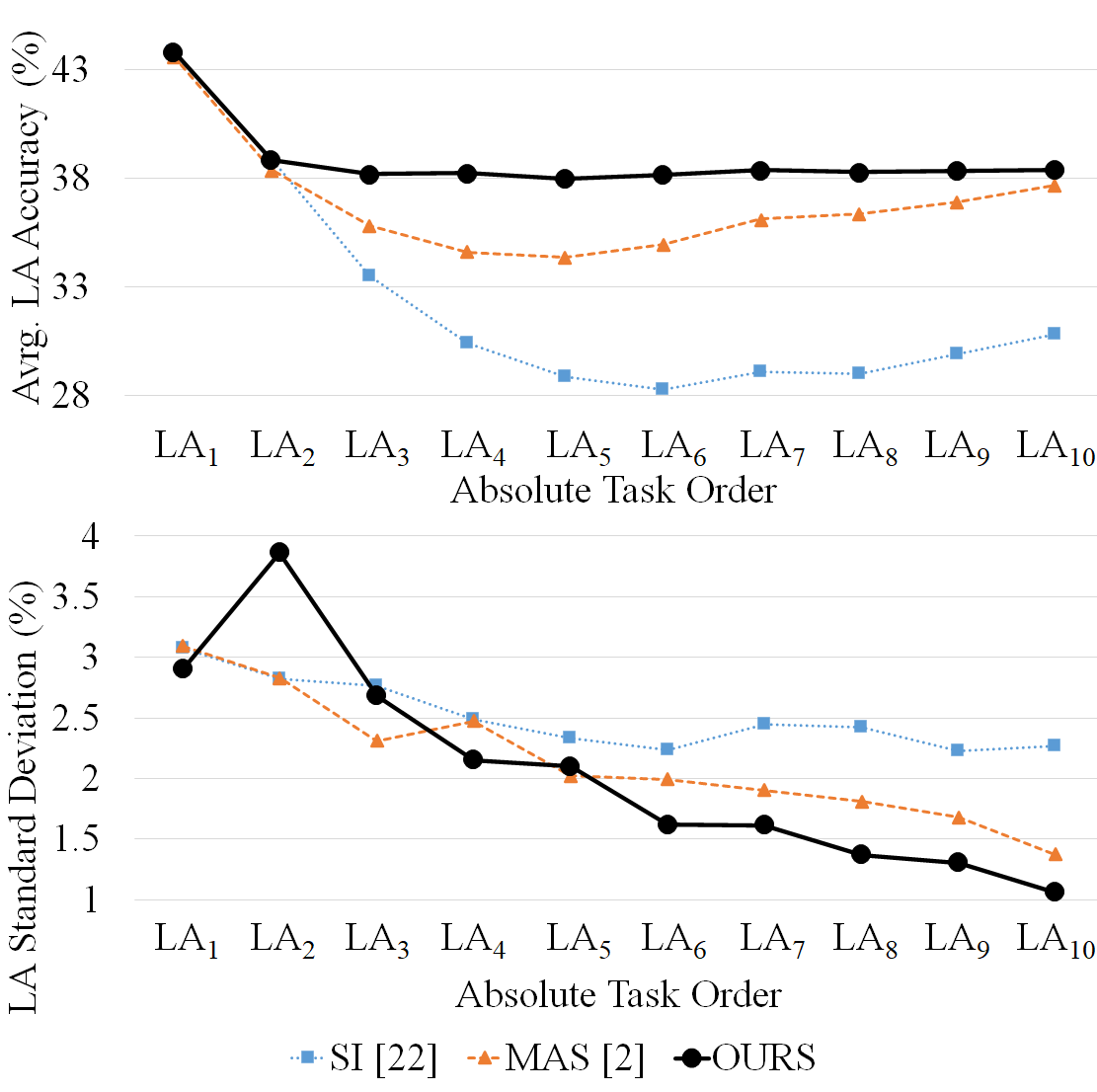}
\caption{Average LA Accuracy and its std. of Tiny ImageNet data set}
\label{tiny}
\end{center}
\end{figure}

\begin{table*}[t!]
\centering
\resizebox{\linewidth}{!}{%
\begin{tabular}{ccccccccccc}
\toprule
\multicolumn{1}{c}{}&\multicolumn{9}{c}{ Average DOI of Absolute Task Order } \\ \cmidrule {2-10}
Method               & D\textsubscript{1}               & D\textsubscript{2}              & D\textsubscript{3}                & D\textsubscript{4}             & D\textsubscript{5}                & D\textsubscript{6}              & D\textsubscript{7}                & D\textsubscript{8}             & D\textsubscript{9}        \\ \midrule
SI~\cite{zenke2017continual}            & 27.35(\textpm3.7)          & 29.02(\textpm4.4)          & 27.09(\textpm4.9)        & 22.67(\textpm4.2)      & 20.24(\textpm3.7)         & 18.25(\textpm4.4)            & 14.86(\textpm3.8)       & 11.70(\textpm3.2)          & 7.78(\textpm3.0)          \\
MAS~\cite{aljundi2018memory}           & 20.49(\textpm4.2)           & 16.32(\textpm3.6)            & 14.40(\textpm4.1)          & 11.02(\textpm3.6)      & 9.14(\textpm3.6)          & 8.27(\textpm2.7)             & 6.20(\textpm2.3)          & 5.41(\textpm2.3)          & 3.86(\textpm1.8)       \\
\textbf{OURS} & \textbf{11.47(\textpm3.6)} & \textbf{6.75(\textpm2.8)} & \textbf{5.78(\textpm2.3)} & \textbf{3.98(\textpm1.5)} & \textbf{3.16(\textpm1.2)} & \textbf{2.76(\textpm1.5)} & \textbf{2.15(\textpm1.1)} & \textbf{1.70(\textpm0.9)} & \textbf{0.98(\textpm0.7)} \\ \bottomrule
\end{tabular}
}
\caption{Average DOI(Degree of interference) and its std.(\%) on Split Tiny ImageNet. Note that proposed method forgets past tasks less regardless of the order of tasks.}
\label{tiny_forget}
\end{table*}

\subsection{Split Tiny ImageNet}

We evaluate our method on Split Tiny ImageNet data set where each task has 20 consecutive classes (total 10 tasks). We use the same multi-headed setup as in the case of Split CIFAR10 and Split CIFAR10-100. We train our network for 100 epochs with $\alpha = 0.5$. The order of tasks is randomly shuffled (total 50 random orders).
Only convolution neural networks based methods are tested for a fair comparison.

In Figure~\ref{tiny}, our method outperforms all other methods with large margins. The standard deviation graph shows that our method algorithm shows the least performance disparity under the change in the order of tasks.
Table~\ref{tiny_forget} presents that our method acquires lowest average degraded degree of performance among SI~\cite{zenke2017continual}, MAS~\cite{aljundi2018memory} and ours, achieving $19.08\%$, $10.5\%$, and $4.3\%$ respectively. Also, ours has the lowest standard deviation of DOI. This implies that our method is robust to the interference from various combinations of tasks.

\section{Conclusion}
We have proposed an activation importance-based continual learning method that consolidates important neurons of past tasks. Comprehensive evaluation has proved that the proposed method has implemented regularization-based continual learning achieving the fundamental aim of continual learning tasks not only balancing between stability and plasticity but also keeping robustness of the performance to the changes in the order of tasks.


{\small
\bibliographystyle{ieee_fullname}
\bibliography{egbib}
}

\end{document}